\documentclass[11pt]{article}

\usepackage{amsmath, amssymb, amscd, amsthm, amsfonts}
\usepackage{graphicx}
\usepackage{hyperref}
\usepackage{listings}
\usepackage{color}
\usepackage{caption}
\usepackage{float}
\usepackage{lipsum}

\usepackage[
style=ieee
]{biblatex}
\usepackage[utf8]{inputenc}
\usepackage{xcolor}
\definecolor{codegreen}{rgb}{0,0.6,0}
\definecolor{codegray}{rgb}{0.5,0.5,0.5}
\definecolor{codepurple}{rgb}{0.58,0,0.82}
\definecolor{backcolour}{rgb}{0.95,0.95,0.92}
\lstdefinestyle{mystyle}{
    backgroundcolor=\color{backcolour},   
    commentstyle=\color{codegreen},
    keywordstyle=\color{magenta},
    numberstyle=\tiny\color{codegray},
    stringstyle=\color{codepurple},
    basicstyle=\ttfamily\footnotesize,
    breakatwhitespace=false,         
    breaklines=true,                 
    captionpos=b,                    
    keepspaces=true,                 
    numbers=left,                    
    numbersep=5pt,                  
    showspaces=false,                
    showstringspaces=false,
    showtabs=false,                  
    tabsize=2
}

\title{\textbf{\huge{Analysis of Evolutionary Program Synthesis for Card Games}}}
\author{\Large{Rohan Saha$^{*1}$, Cassidy Pirlot$^{*2}$} \\ \{rsaha$^1$, pirlot$^2$ \}@ualberta.ca}
\date{\today}

\newcommand\blfootnote[1]{%
  \begingroup
  \renewcommand\thefootnote{}\footnote{#1}%
  \addtocounter{footnote}{-1}%
  \endgroup
}

\begin{document}

\maketitle

\begin{abstract}
In this report, we inspect the application of an evolutionary approach to the game of Rack'o, which is a card game revolving around the notion of decision making. We first apply the evolutionary technique for obtaining a set of rules over many generations and then compare them with a script written by a human player. A high level domain specific language is used that determines which the sets of rules are synthesized. We report the results by providing a comprehensive analysis of the set of rules and their implications.
\end{abstract}

\section{Introduction}
\label{section-introduction}
A genetic algorithm is a search heuristic that aims to find optimal solutions through ideas found in biology. This includes concepts such as survival of the fittest, mutations, cross-breeding, and more, in other words, an evolutionary approach.  We previously saw in assignment 1, the performance of such an algorithm in a smaller version of the game CAN'T STOP, here we aim to evaluate the performance of it in the game RACK'O. \blfootnote{* - Both authors have equal contributions. Ordering was decided over a coin toss on a Google Meet.}\par
Evolutionary approach is a method where the goal is to find a solution to a problem iteratively given a initial value. In the context of program synthesis, evolutionary approaches are used to generate a set of rules that is consistent with the game mechanics and this set of rules is expected to perform better than other scripts in the same search space. The set of rules that is obtained using a fitness function that measures how good the set of rules are given a state of the game. We chose to investigate the area of evolutionary approach in program synthesis because it is an interesting method to generate strategies and research using evolutionary approach spans over a multitude of program synthesis problems such as program sketching\cite{Badek2017} and guided search for synthesizing programs with high complexity\cite{Sasanka2019}. \par As a overview, we first explain the game and its mechanics in section \ref{section:game_racko}, the structure of the project in \ref{section:project_structure}, the domain specific language in \ref{section:dsl}, the design of the experiment in section \ref{section:experiment_design}, the results and findings in section \ref{section:results_and_discussion}, and finally conclude in section \ref{section:conclusion}.

\section{Game: RACK'O}
\label{section:game_racko}
Rack'O is a card game where the objective for a player is to arrange the cards in your hand in ascending order. The number of players can range from two to four. The game has two decks of cards, a discard pile and a Rack'o deck. Initially, all the cards are shuffled randomly to create the Rack'o deck. Then each player is dealt ten cards from the Rack'o deck, which is known as the player's hand. The discard pile is started by turning over the top card from the Rack'o deck. Each player, during their turn, picks up a card from the Rack'o deck, and decides to either replace a card from their hand or does not. The player can also pick up the card from the discard pile and decide to whether replace a card in hand or not. This means that it may be possible that the player decides to pass. \par In the project, a smaller version of the game is used for simplicity. Forty total cards are used instead of sixty and each player is dealt five cards instead of ten. Also, the game is designed for two players instead of a maximum of four. These changes are considered due to computational restrictions, but the project can easily be expanded to the full version of the game.

\section{Project Structure}
\label{section:project_structure}
\begin{itemize}

    \item  \textbf{DSL.py-} This file contains our domain specific language as well as our context free grammar which uses \texttt{if's}, \texttt{and's}, and the names of the functions from our DSL.
    \item \textbf{Evaluation.py-} This file has all of the functions necessary to preform the genetic algorithm. This includes:
    \begin{itemize}
    \item \textbf{Evaluation} This function plays takes in two instantiated scripts and has them play Rack'o against each other n times and returns the amount of times each player won as well as the relative rate of winning for each payer (the amount of times they won over n for each player).
    \item \textbf{Eval} This takes in a dictionary of scripts as the keys with their corresponding fitness scores as the values as well as an integer m.  It causes each script in the dictionary to play in the Evaluation function, as both player one and two, m times and return the updated dictionary.
    \item \textbf{Elite} This takes in a dictionary scripts with their fitness scores as values and an integer k.  This function compares the fitness scores of the scripts and returns a reduced dictionary of the best k scripts according to the fitness scores.
    \item \textbf{Tournament} This function takes in the dictionary as well as an integer t. It randomly takes t scripts from the dictionary and returns the best two scripts according to the fitness score.
    \item \textbf{GenerateSplit} Takes in two 'parent' scripts and returns a random crossover strategy in a string.
    \item \textbf{crossover} This function creates and returns the script that the GenerateSplit function returns.
    \item \textbf{mutate} Takes in a script mutates, instantiates, and returns it.
    \item \textbf{RemovedUnused} This function takes in a single script and looks at if a certain rule that the script contains has ever been used. If it has not then it removes this rule. The adjusted script is returned.
    \item \textbf{EZS} This is our genetic algorithm which takes in a generation number, population size, the number of desired elites, and the size of the tournament. This returns our generated script.
    \end{itemize}
    \item \textbf{Script.py-} Contains the outline in which our scripts get generated as.
    \item \textbf{OurScript.py-} This is the script we made ourselves (more on this later).
    \item \textbf{Game.py-} This contains functions for how the game is played including shuffling, dealing, seeing available moves that can be made, etc. These functions are used in the Evaluation function in Evaluation.py.
\end{itemize}

\section{DSL}
\label{section:dsl}
The domain specific language(DSL) represents the set of template functions or methods that define the rules a player can play during the game.
The following rules were included in the DSL which were used to generate a script.\\

The parameters used are as follows:
\begin{itemize}
    \item \textbf{action}: the resulting hand after a play is made on the players hand.
    \item \textbf{index}: represents the index in a players hand that we are currently concerned with.
    \item \textbf{hand}: the current hand of a player
    \item \textbf{number}: possible number from the Rack'o deck (0-39)
\end{itemize}

\begin{enumerate}
    \item \textbf{isBigger(action, index, hand)}: Returns true if the card picked up from either the Rack'o deck or discard pile, when swapped with a card at the index in the player's hand, is bigger than the card below it, else returns false.
    \item \textbf{isSmaller(action, index, hand)}: Returns true if the card picked up from either the Rack'o deck or discard pile, when swapped with a card at the index in the player's hand, is smaller than the card below it, else returns false.
    \item \textbf{givesRacko(action)}: Returns true 
    if the player achieves Rack'o by taking the action.
    \item \textbf{hasRacko(hand)}: Returns true if the player has all cards arranged in ascending order.
    \item \textbf{isCardBetweenNumbers(action, number, number, index, hand)}: Returns true if the card picked from either the Rack'o deck and discard pile, when placed in the index, is in between the two specified numbers.
\end{enumerate}
Using the rules from the DSL, at each generation, a set of random scripts are created initially and then the best performing scripts are passed on to the next generation using various genetic operations such as crossover and mutation.
\section{Experiment Design}
\label{section:experiment_design}

\subsection{Running of Genetic Algorithm}
\label{combinations}
We ran the algorithm multiple times with differing EZS parameters. Each time 100 games were played to compare two scripts and the scripts were compared 3 times as player 1 each.  The different cases are as follows:
\begin{enumerate}
    \item Population: 10, Generations: 4, Elites: 7, Tournaments: 5
    \item Population: 20, Generations: 6, Elites: 7, Tournaments: 7
    \item Population: 30, Generations: 8, Elites: 10, Tournaments: 10
\end{enumerate}
\subsection{Playing against our own script}
To test how well the genetic algorithm performed we consider testing it against a script that mimics how we would play the game given the DSL. We play by placing the drawn card in a chosen slot if it fits in the corresponding interval for that slot.  The (exhaustive) intervals are all of equal length and contain larger values for the higher index slots.  The code is found below:
\lstset{style=mystyle}
\begin{lstlisting}[language=Python, firstline = 0, lastline =55]
from Player import Player
import random
from Game import Game
from DSL import DSL


class OurScript(Player):

    def __init__(self):
        self._id = 1234567
        self._strategies = ['DSL.givesRacko(a)',
                        'DSL.isCardBetweenNumbers(a,1,8,0,Game.getRack())',
                        'DSL.isCardBetweenNumbers(a,9,16,1,Game.getRack())',
                        'DSL.isCardBetweenNumbers(a,17,25,2,Game.getRack())',
                        'DSL.isCardBetweenNumbers(a,26,34,3,Game.getRack())',
                        'DSL.isCardBetweenNumbers(a,35,40,4,Game.getRack())']
        self._counter_calls = []
        for i in range(5):
            self._counter_calls.append(0)

    def get_counter_calls(self):
        return self._counter_calls

    def get_action(self, Game, card, hand):
        actions = Game.available_moves(card, hand)


        for a in actions:


            if DSL.givesRacko(a):
                self._counter_calls[0] += 1
                return a

            if DSL.isCardBetweenNumbers( a, 1 , 8 , 0 , Game.getRack()):
                self._counter_calls[0] += 1
                return a

            if DSL.isCardBetweenNumbers( a, 9 , 16 , 1, Game.getRack()):
                self._counter_calls[1] += 1
                return a

            if DSL.isCardBetweenNumbers( a, 17 , 25 , 2 , Game.getRack()):
                self._counter_calls[2] += 1
                return a

            if DSL.isCardBetweenNumbers( a, 26 , 34 , 3 , Game.getRack()):
                self._counter_calls[3] += 1
                return a

            if DSL.isCardBetweenNumbers( a, 35 , 40 , 4 ,Game.getRack() ):
                self._counter_calls[4] += 1
                return a

        return actions[0]
\end{lstlisting}

We tested our script against the generated script with the evaluation function found in Evaluation.py letting them alternate being player one in case there is some sort of advantage in being in that position.

\section{Results and Discussion}
\subsection{Resulting Scripts}
\label{section:results_and_discussion}
For case 1 the simulation ran for 10-15 minutes the resulting script had rules:
\begin{itemize}
    \item `DSL.givesRacko(a)'
    \item `DSL.isCardBetweenNumbers(a, 37 , 39 , 3 )'
    \item `DSL.isCardBetweenNumbers(a, 25 , 29 , 0 )'
    \item `DSL.hasRacko(Game.getRack())' 
    \item `DSL.isCardBetweenNumbers(a, 34 , 20 , 4 )' \item `DSL.isSmaller(a, 1 , Game.getRack() )' \item `DSL.isSmaller(a, 2 , Game.getRack() )'
\end{itemize}
Case 2 ran for just over an hour and resulted in a script with:
\begin{itemize}
    \item `DSL.isSmaller(a, 0 , Game.getRack() )'
    \item `DSL.isCardBetweenNumbers(a, 39 , 26 , 0 )'
    \item `DSL.isBigger(a, 3 , Game.getRack() )'
    \item `DSL.isSmaller(a, 4 , Game.getRack() )' \item `DSL.isCardBetweenNumbers(a, 9 , 13 , 0 )' \item `DSL.isSmaller(a, 3 , Game.getRack() )' \item `DSL.isCardBetweenNumbers(a, 37 , 18 , 3 )' \item `DSL.isCardBetweenNumbers(a, 33 , 8 , 4 )' \item `DSL.isCardBetweenNumbers(a, 3 , 31 , 1 )' \item `DSL.isCardBetweenNumbers(a, 15 , 37 , 3 )' \item `DSL.isBigger(a, 0 , Game.getRack() )' \item `DSL.isBigger(a, 1 , Game.getRack() )' \item `DSL.hasRacko(Game.getRack())'
    \item `DSL.isSmaller(a, 2 , Game.getRack() )' \item `DSL.isBigger(a, 2  Game.getRack() )' 
    \item `DSL.givesRacko(a)' 
    \item `DSL.isCardBetweenNumbers(a, 9 , 28 , 0 )'
\end{itemize}
Case 3 ran for just over 2 hours and resulted in a script with:
\begin{itemize}
    \item `DSL.isBigger(a, 2 , Game.getRack() )'
    \item `DSL.givesRacko(a)'
    \item `DSL.hasRacko(Game.getRack())'
    \item `DSL.isCardBetweenNumbers(a, 21 , 12 , 3 )'
    \item `DSL.isCardBetweenNumbers(a, 26 , 33 , 3 )'
    \item `DSL.isCardBetweenNumbers(a, 4 , 34 , 4 )'
    \item `DSL.isBigger(a, 1 , Game.getRack() )'
    \item `DSL.isSmaller(a, 3 , Game.getRack() )'
    \item `DSL.isCardBetweenNumbers(a, 12 , 3 , 2 )'
\end{itemize}

\subsection{Comparison to Our Script}
We played Rack'o with the resulting scripts from each of the three cases against our script that we wrote. First of all, it must be noted that though the DSL may seem limited, the search space is actually very large because the parameters of the functions can have a wide range of values. This makes it difficult for the evolutionary approach to find a script that is globally optimal. It is often necessary to have an extremely large population with many generations to obtain such an optimal script but due to computational restrictions, we only run the algorithm for the three cases explained in section \ref{combinations}.\par  Let's now have a look the rules generated from the evolutionary approach. We can see that for each case, the rule, givesRacko(a) is present. This indicates that it is advisable to play an action if it helps to achieve Rack'o. This is fairly obvious and is an easy strategy to follow. The other rule that is present in the best script for all the three cases is the isCardBetweenNumbers(...) function. This is a good strategy because the knowledge of the cards in the current hand is being used to check whether the newly swapped card will help to have all cards in ascending order. We found that the other strategies present do not contribute as much to performance as this will be evident while observing the results given below.

\begin{figure}[H]
  \centering
    \includegraphics[width=8cm]{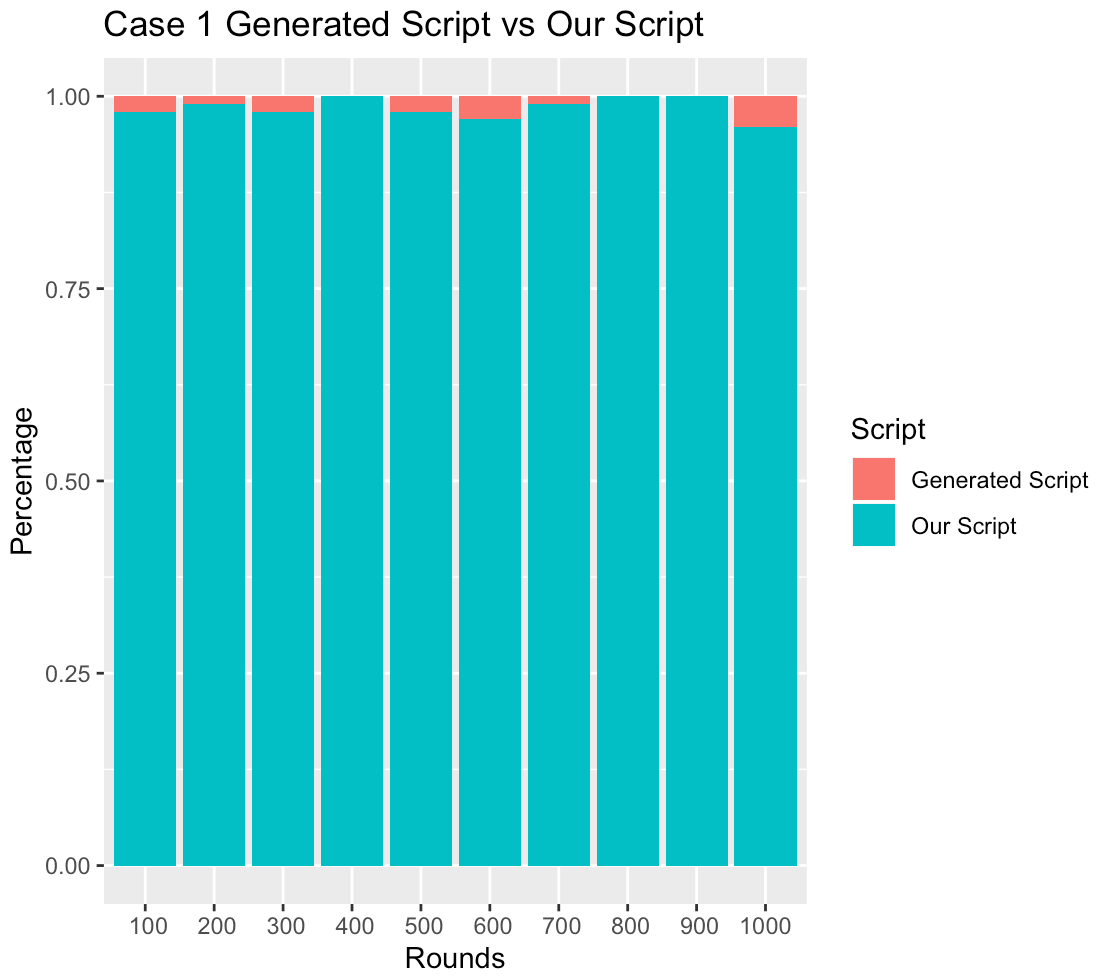}
    \centering
    \includegraphics[width=8cm]{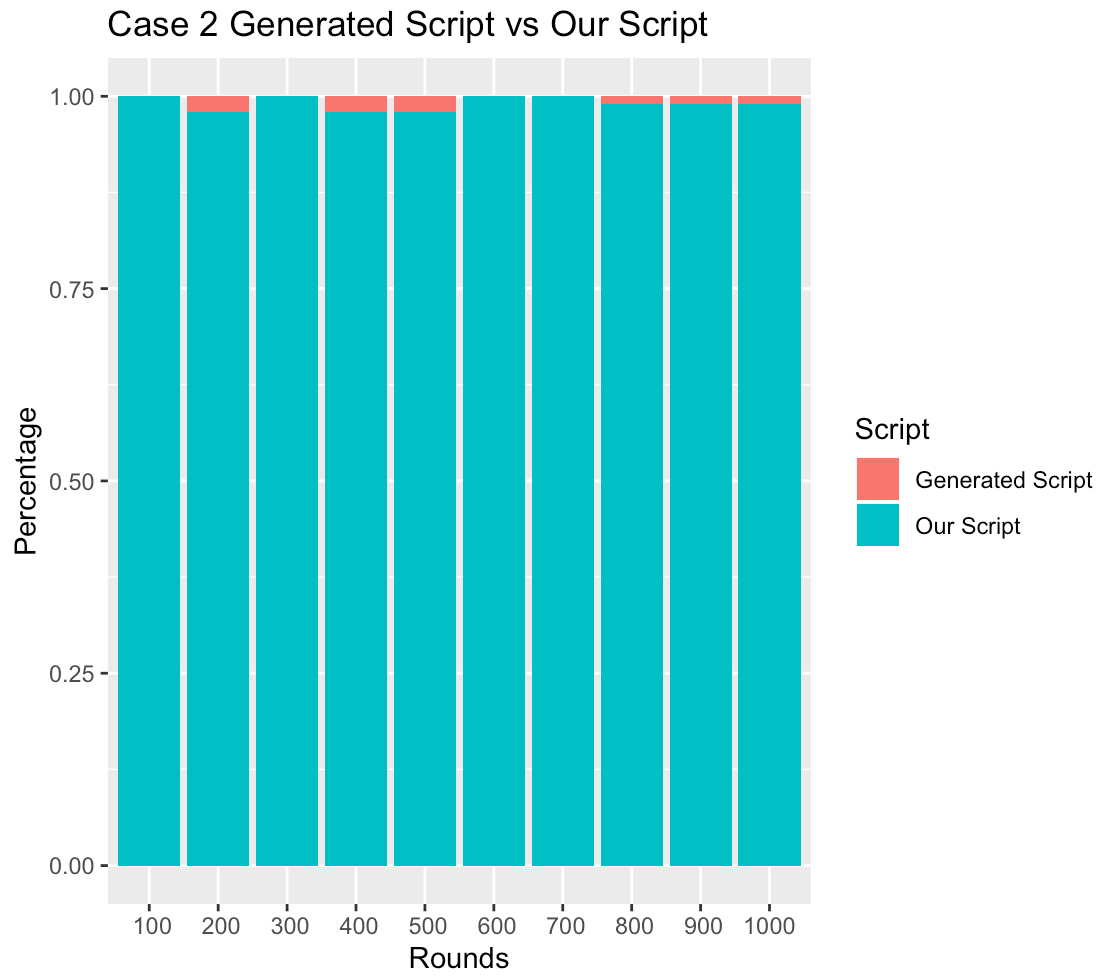}
    \centering
    \includegraphics[width=8cm]{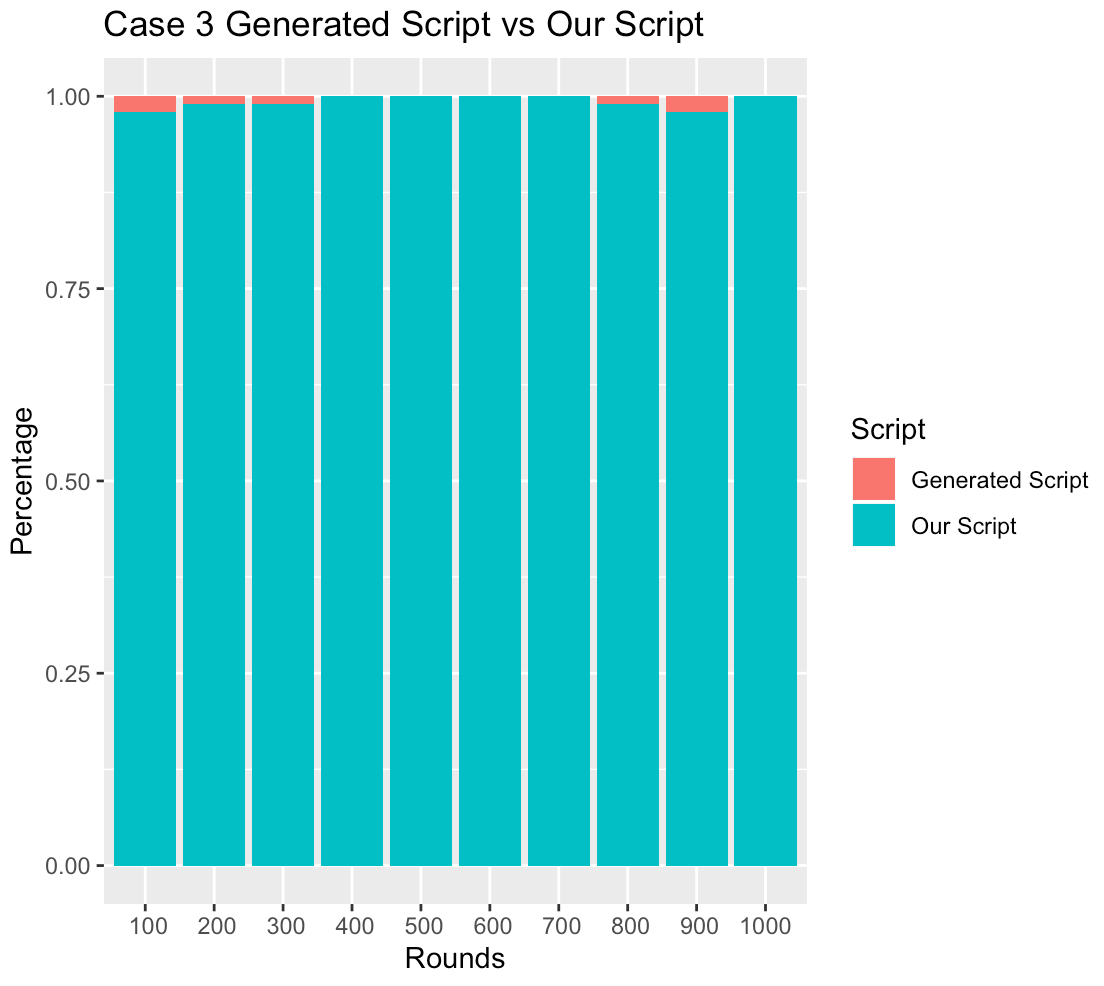}
    \caption{Percentage wins of the best script from the Evolutionary algorithm vs OurScript for all cases}
\end{figure}

We see that the generated scripts did not perform very well against our script. However, we are impressed that they were able to win any games at all. We were expecting that as we increased our population size and generations that there would be improvement against our script, which we only see marginally. This may be because there was one strategy that was exceptionally stronger than the others (isBetweenNumbers) and our script was made entirely of that one function. \\ All in all, from the results and the generated scripts, the functions givesRacko(...) and isCardBetweenNumbers(...), where the ellipses are the arguments, are better than the other functions in the DSL. This is because preference is given to the actions that help to achieve Rack'o and to cards when picked up from the either the discard pile or Rack'o deck and swapped that may help to achieve Rack'o in the future.

\section{Conclusion}
\label{section:conclusion}
In this project, we investigated the applicability of the evolutionary approach with respect to program synthesis. Unlike assignment 1, we looked at how an evolutionary approach would be feasible in the domain of card games. We used a simplified version of Rack'o and analyzed the rules generated from the evolutionary algorithm. \par We ran the evolutionary algorithm for different population sizes and compared the generated rules with a custom script containing the functions from the DSL. We found that our own script outperformed the scripts generated from the EZS almost all the time. This was because our script was carefully tailored to contain specific arguments to the strategies and thus facilitated the arrangement of cards in ascending order. On the other hand, the scripts generated from the evolutionary approach had more stochasticity in terms of the values of the parameters. Overall, it can be observed that even if the functions are same, the likelihood of achieving Rack'o is highly pertinent on the cards in the current hand and the card selected from the Rack'o deck or the discard pile. In addition, from such observations, it is advisable to understand the game mechanics before formulating a strategy, which will elevate the chances of winning the game.\par
In the future, stronger functions in the DSL would be used and even larger population sizes and generations should be used as well as increasing the amount of games played during each evaluation.  We suspect that because our DSL has a large space, due to having to include all numbers 0-39, that it would take longer for our scripts to explore and thus converge to an optimal script.\par Evaluating such computational methods for synthesizing scripts help to understand the game mechanics better and devise better strategies that would contribute to the evolution of better and more sophisticated card games.

\section{Code}
\label{section:code}
The code for the project is given in the following GitHub repository: \\ \href{https://github.com/simpleParadox/CMPUT-659-Project}{https://github.com/simpleParadox/CMPUT-659-Project}.

\printbibliography

\end{document}